\documentclass[letterpaper, 10 pt, conference]{ieeeconf}  
\IEEEoverridecommandlockouts                              

\usepackage{graphics} 
\usepackage{epsfig} 
\usepackage{times} 
\usepackage{amsmath} 
\usepackage{amssymb}  
\usepackage{cite}
\usepackage{siunitx} 
\usepackage{soul}
\usepackage{color}
\usepackage{xcolor}

\soulregister\ref7  
\soulregister\cite7 
\soulregister\eqref7
\soulregister\prettyref7

\usepackage{multirow}
\usepackage{algorithm} 
\usepackage{algpseudocode}
\usepackage{makecell}
\usepackage{booktabs}
\usepackage{threeparttable}
\usepackage{adjustbox}

\bstctlcite{IEEEexample:BSTcontrol}

\usepackage{prettyref}
\newrefformat{fig}{Fig.~\ref{#1}}
\newrefformat{sec}{Sec.~\ref{#1}}
\newrefformat{tab}{Table~\ref{#1}}
\newrefformat{alg}{Algo.~\ref{#1}}
\newrefformat{eq}{Eq.~(\ref{#1})}

\title{\LARGE \bf
RotateIt! Fast and Reliable Single-Arm Garment Unfolding via Online-Adaptive Dynamic Rotation
}
\author{
Zeqing Zhang$^{1,*}$, Zuokun Xie$^{1,*}$, Ao Fang$^{1}$, Bin Dai$^{1}$, Zhengjie Shu$^{2}$, Yifeng Tang$^{3}$, Ziwei Wang$^{1,\dagger}$
\thanks{$^*$ Equal Contributions. $^\dagger$ Corresponding Author.}
\thanks{$^{1}$ Nanyang Technological University.}
\thanks{$^{2}$ The University of Hong Kong.}
\thanks{$^{3}$ Chinese Academy of Sciences‌.}
}

\begin{document}
\maketitle
\thispagestyle{empty}
\pagestyle{empty}


\begin{abstract}

Robotic garment unfolding is essential for downstream tasks, yet quasi-static methods require repeated actions, while existing dynamic approaches predominantly rely on bimanual flinging. We present \textit{RotateIt!}, a single-arm framework that uses adaptive axial rotation for dynamic garment unfolding. To the best of our knowledge, it is the first unfolding framework to employ dynamic axial rotation as its primary manipulation primitive. From a randomly initialized tabletop configuration, the robot selects a rotation-effective grasp and rotates the lifted garment about an approximately fixed anchor, generating inertial tension that separates overlapping layers within a compact workspace. A grasp ranker selects the anchor, while an online residual policy adapts the rotation extent and speed, thereby determining the release timing. Across seen and unseen simulated garments and eight unseen real garments, \textit{RotateIt!} improves success within three attempts by $44.0$--$61.0$ percentage points over quasi-static pick-and-place. The simulation-trained policies transfer zero-shot to the real world, achieving $75.6\%$ success, $41\%$ higher first-attempt coverage, and $26\%$ higher final coverage. The resulting states further enable autonomous robotic folding without manual rearrangement.
\end{abstract}

\section{Introduction}
\label{sec:intro}
Deformable object manipulation (DOM) supports applications ranging from
domestic assistance and healthcare to automated textile manufacturing.
Garments are particularly challenging due to their high-dimensional
configurations, severe self-occlusion, complex contact dynamics, and
diverse geometries and material properties. Despite substantial progress,
efficiently transforming a randomly configured garment into a sufficiently
unfolded state remains a fundamental bottleneck for downstream tasks such as folding and ironing \cite{longhini2025unfolding}.

Extensive prior work addresses garment unfolding using quasi-static
primitives, including pick-and-place, pick-and-drag, and repeated
regrasping
\cite{wu2019learning,lin2022learning,weng2022fabricflownet}.
Although reliable and straightforward to execute, these actions typically
produce localized deformations and require repeated perception-action
cycles to achieve high coverage. The resulting execution time and
accumulated perception and grasping errors are restrictive when
unfolding serves only as an initialization stage for a longer
garment-manipulation pipeline.

\begin{figure}[t]
    \centering
    \includegraphics[width=\linewidth]{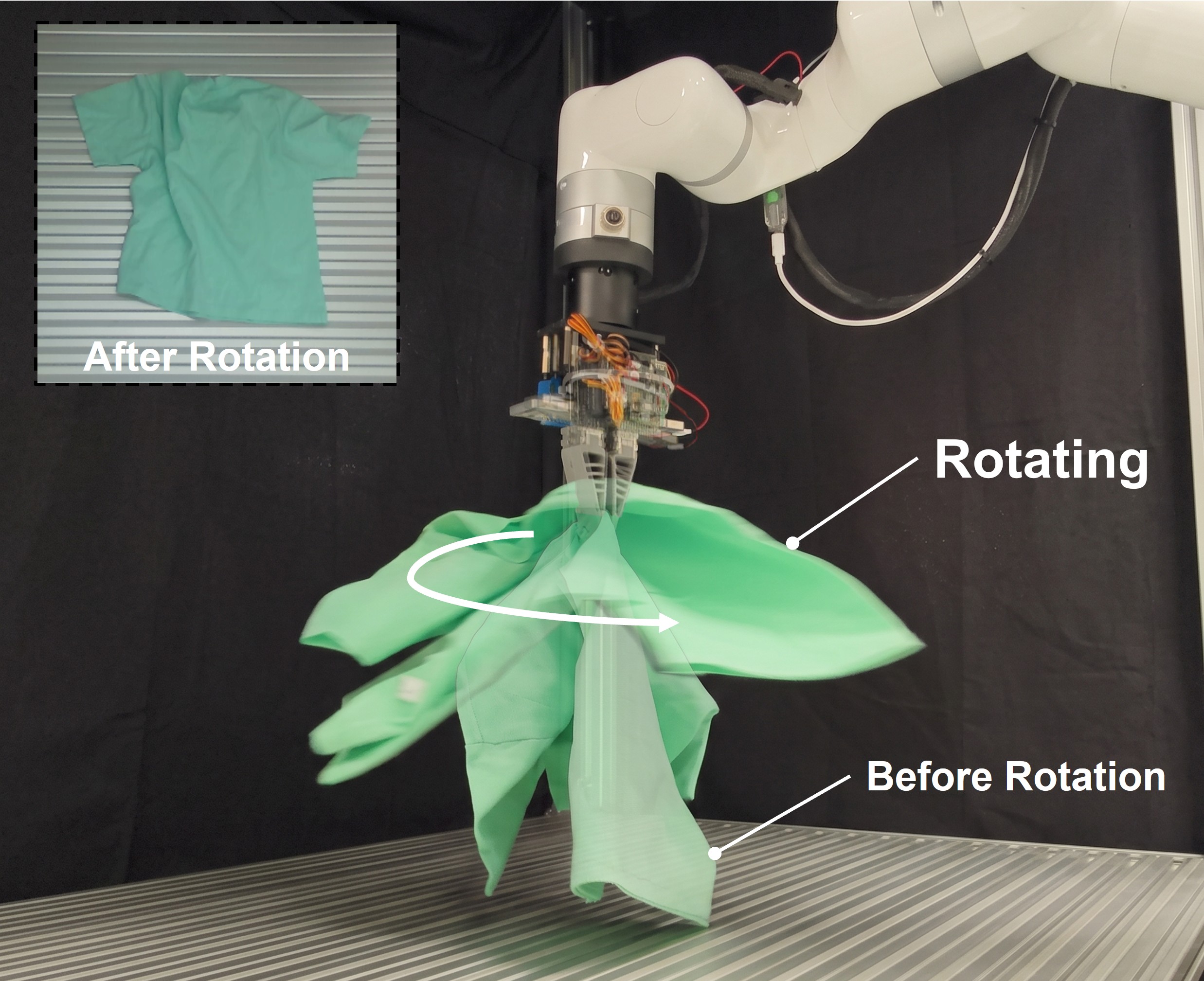}
    \vspace{-20pt}
    \caption{The proposed \textit{RotateIt!} achieves fast, high-coverage cloth unfolding through adaptive rotation and release, transferring zero-shot from simulation to unseen real garments. Overlaid frames show axial rotation; the inset shows the settled result after release.}
    \label{fig:real_hardware}
    \vspace{-10pt}
\end{figure}

Dynamic garment manipulation (DGM) instead exploits garment inertia to
produce large, nonlocal deformations with fewer interactions. Existing
systems have demonstrated effective unfolding and goal-conditioned
manipulation through bimanual flinging and its combination with
quasi-static refinement
\cite{ha2022flingbot,canberk2023cloth,yang2024one}.
However, these methods require two suitable grasp points, coordinated
dual-arm motion, and sufficient workspace for stretching and flinging,
limiting their applicability to common single-arm platforms.

Single-arm DGM remains underexplored. Existing work
optimizes translational fling trajectories from predefined,
garment-dependent grasps and may require additional shaking or real-world
adaptation for novel garments \cite{chen2022efficiently}. Rotational
fabric manipulation has also been studied for maintaining periodic motion
\cite{liu2026periodic}, but not for producing a terminal unfolded state.
Thus, existing dynamic unfolding remains centered on translational
flinging, while dynamic axial rotation has not been exploited as a
garment-unfolding primitive.

These limitations motivate a fundamental question:
\textit{Can dynamic rotational manipulation serve as an effective
primitive for garment unfolding?} We answer this question with
\textbf{\textit{RotateIt!}}, a single-arm dynamic garment-unfolding framework
based on adaptive axial rotation.
Starting from a randomly initialized tabletop configuration, the robot selects one
grasp, lifts the garment, and rotates it about an approximately fixed
anchor, as illustrated in \prettyref{fig:real_hardware}. The resulting rotation-induced inertial tension separates
overlapping layers and expands the suspended garment within a compact
workspace. To control this dynamic process, \textit{RotateIt!} addresses
four coupled decisions: \textbf{where to grasp, how fast to rotate, how
far to rotate, and when to release}. A grasp-ranking model selects a
rotation-effective grasp, while an observation-conditioned residual policy
adapts the rotation speed and extent online, thereby determining the
release timing. Both policies are trained entirely in simulation and
deployed on unseen real garments without real-world fine-tuning.
The experiments show fast, high-coverage, and reliable unfolding results.

To the best of our knowledge, \textit{RotateIt!} is the first garment-unfolding framework to employ dynamic axial rotation as its primary manipulation primitive. \textbf{Main Contributions:}
\begin{itemize}
    \item We introduce \textit{RotateIt!}, a single-arm dynamic garment-unfolding framework via adaptive axial rotation to generate sustained rotation-induced inertial tension.

    \item We formulate garment unfolding as a coupled grasp-and-rotation decision problem and solve it with a two-stage policy that conditions online rotate-and-release control on the selected grasp and evolving garment state.

    \item We evaluate \textit{RotateIt!} in simulation and the real world, showing efficient and generalizable unfolding, zero-shot sim-to-real transfer, and applicability to downstream tasks.

    \item We provide an open-source benchmark for single-arm dynamic garment unfolding, including simulation environments, garment models, randomized initial states, trained policies, and baseline implementations.
\end{itemize}

The remainder of this paper is organized as follows. 
\prettyref{sec:related} reviews related work and \prettyref{sec:method} introduces the proposed grasp-selection and adaptive rotate-and-release policies. \prettyref{sec:exp} reports the simulated and real-world experimental results, including zero-shot sim-to-real transfer and downstream folding. Finally, \prettyref{sec:conclusion} concludes the paper and discusses future directions.

\begin{figure*}[t]
    \centering
    \includegraphics[width=0.95\textwidth]{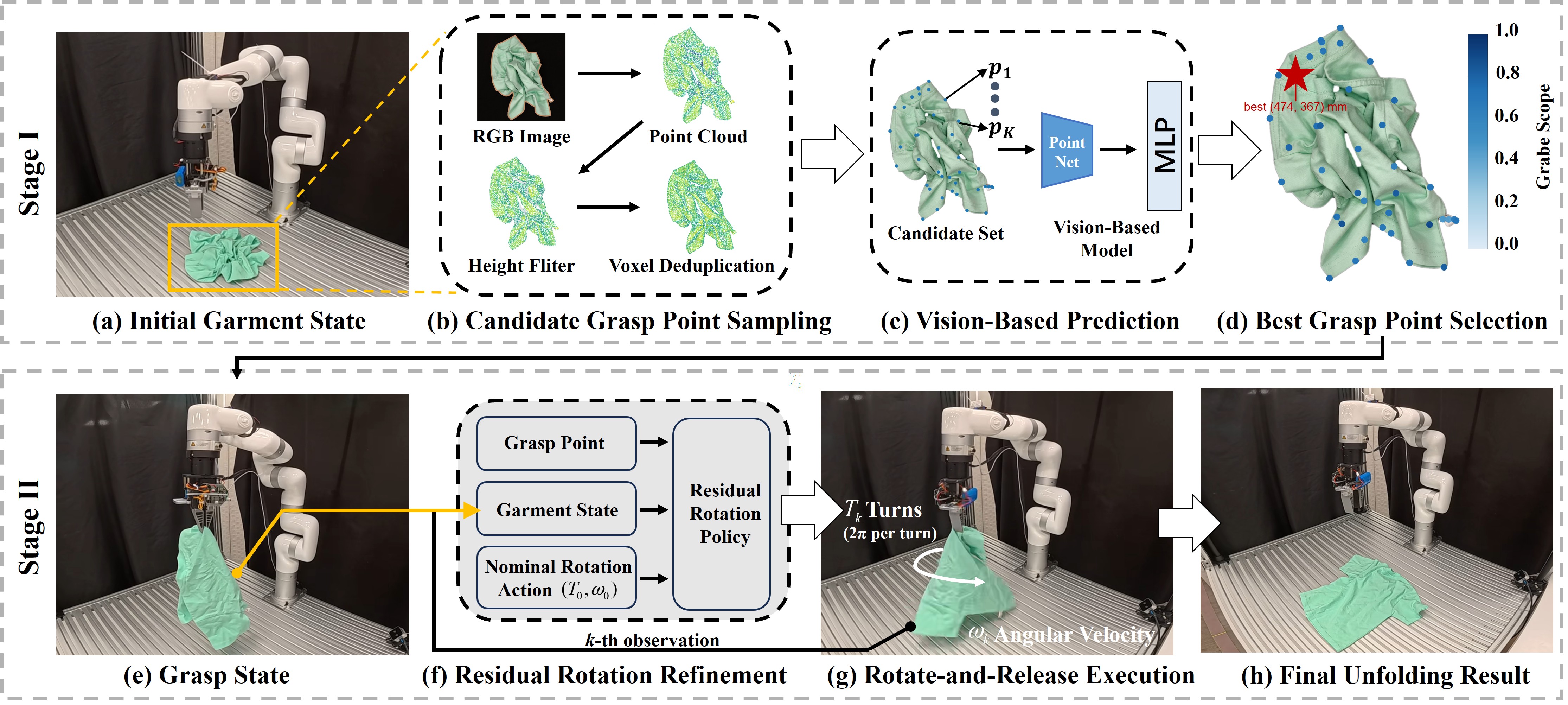}
    \vspace{-10pt}
    \caption{Overview of \textit{RotateIt!}. Stage I processes the initial RGB-D observation, samples and scores visible grasp candidates, and selects a rotation-effective grasp (a)--(d). After grasping and lifting (e), Stage II uses online observations of the suspended garment to generate residual adjustments to the nominal rotation extent and angular velocity over successive motion chunks (f)--(g). The resulting adaptive rotation schedule determines the release timing, after which the garment settles into an unfolded state (h). Two RGB-D cameras are positioned at the upper-left and lower-right of the current view; owing to cropping for visual clarity, they are not visible in all panels.}
    \label{fig:pipeline_overview}
    \vspace{-20pt}
\end{figure*}
\section{Related Work}
\label{sec:related}

\subsection{Quasi-Static Garment Unfolding}
Early garment-unfolding systems primarily relied on geometric reasoning and sequential regrasping. Geometric methods detected garment boundaries, corners, or hem features and used a sequence of hanging observations and regrasps to recover a spread configuration \cite{triantafyllou2016geometric}. Predictive models were later introduced to evaluate candidate grasps and iteratively transform garments from unknown configurations into recognizable states \cite{li2015regrasping}. Although these methods provide interpretable manipulation strategies, they require multiple perception, grasping, and rehandling steps.
Recent learning-based approaches select quasi-static pick-and-place or pick-and-drag actions using self-supervision, learned dynamics, optical flow, or spatial action maps \cite{wu2019learning,lin2022learning,weng2022fabricflownet,lee2021learning}. While these methods enable precise, goal-directed deformation, unfolding is achieved through successive, surface-mediated actions. Severely overlapped garments may therefore require repeated perception-action cycles, increasing execution time and compounding grasping and prediction errors. This motivates dynamic primitives that induce larger configuration changes with fewer interactions.


\subsection{Dynamic Garment Manipulation}


Dynamic garment manipulation exploits garment inertia to induce large,
nonlocal deformations. FlingBot learns bimanual grasp locations for a
scripted stretch--fling--place routine \cite{ha2022flingbot}, while Cloth
Funnels combines dynamic flinging with quasi-static refinement for garment
canonicalization \cite{canberk2023cloth}. One Fling to Goal further uses
environment-aware dynamics to refine bimanual fling trajectories online
for goal-conditioned placement \cite{yang2024one}. Despite their different
control formulations, these methods require 
coordinated bimanual motion and sufficient workspace for stretching and
flinging.

Single-arm DGM remains less explored. Chen et al.\ optimize parameterized
fling trajectories from manually specified, garment-dependent grasps and
adapt to novel garments through real-world trials
\cite{chen2022efficiently}. Dynamic actions have also been studied for
specialized tasks, including bimanual gown unfolding
\cite{blanco2025evaluating} and periodic fabric spinning
\cite{liu2026periodic}. Although periodic spinning demonstrates that
rotation can extend flexible fabric, it targets sustained motion rather
than terminal unfolding through adaptive rotation and release. Existing
DGM therefore remains centered on translational flinging or sustained
spinning, rather than employing dynamic axial rotation as an unfolding
primitive. In contrast, \textit{RotateIt!} uses approximately fixed-anchor
rotation to unfold garments within a compact single-arm workspace.

\subsection{Grasp-Conditioned Dynamic Control}

Selecting an effective grasp is fundamental to garment manipulation because it determines both the attachment point and the suspended garment geometry after lifting. Classical methods detect corners and boundaries using geometric cues \cite{maitin2010cloth}, while learning-based methods estimate grasp affordances from RGB-D observations or point clouds \cite{de2026dataset}. Category-level correspondence can further transfer semantic grasp locations across related garments \cite{wu2024unigarmentmanip}. However, a visually or semantically salient point is not necessarily an effective anchor for dynamic rotation.
Existing methods typically optimize one side of the grasp--motion coupling while holding the other fixed. FlingBot learns bimanual grasps for a predefined dynamic routine \cite{ha2022flingbot}, whereas single-arm flinging optimizes motion parameters from a manually specified grasp \cite{chen2022efficiently}. One Fling to Goal performs online trajectory refinement, but assumes that the fabric is already held at two predefined attachment points \cite{yang2024one}. These formulations do not jointly address autonomous single-arm grasp selection and motion adaptation conditioned on the resulting suspended state.
Residual reinforcement learning can adapt a physically meaningful nominal controller without generating an unconstrained robot trajectory \cite{johannink2019residual}. 
Building on this principle, \textit{RotateIt!} couples rotation-effective
grasp ranking with online residual control, conditioned on the suspended garment state induced by the selected grasp.
\section{Methodology}
\label{sec:method}


\subsection{Problem Statement}
As illustrated in \prettyref{fig:pipeline_overview},
we consider a garment placed in a randomly initialized configuration on a tabletop. Its geometric configuration at time $t$ is represented as
\begin{equation}
    \mathcal{X}_t
    =
    \left\{
        \mathbf{x}_{i,t}\in\mathbb{R}^{3}
    \right\}_{i=1}^{N},
    \label{eq:garment_configuration}
\end{equation}
where $\mathbf{x}_{i,t}$ is the position of the $i$-th garment element. 

Given the initial observation $O_0$, Stage I selects a grasp point
$\mathbf{g}^{\star}$ from the visible candidate set
$\mathcal{G}(O_0)$. During the subsequent rotation, Stage II maps a
compact, grasp-conditioned garment-state representation
$\mathbf{o}_k$ to a bounded residual action $\mathbf{a}_k$:
\begin{equation}
\begin{aligned}
    \text{
    Stage I:
    }
    \:
    \mathbf{g}^{\star}
    &=
    \pi_{\mathrm{I}}(O_0),
    &
    \mathbf{g}^{\star}
    &\in
    \mathcal{G}(O_0),\\
    \text{
    Stage II:
    }
    \:
    \mathbf{a}_k
    &=
    \pi_{\mathrm{II}}(\mathbf{o}_k),
    &
    k
    &=
    0,\ldots,K-1.
\end{aligned}
\label{eq:two_stage_policies}
\end{equation}
Here, $K$ denotes the number of online control chunks in the rotation
phase, and $\mathbf{o}_k$ depends on $\mathbf{g}^{\star}$ through its
grasp-relative descriptors.
%
%
The task-level objective is to maximize the expected coverage
of the settled garment:
\begin{equation}
    \max_{\pi_{\mathrm{I}},\,\pi_{\mathrm{II}}}
    \;
    \mathbb{E}
    \left[
        C\left(
            \mathcal{X}_{\mathrm{final}}
        \right)
    \right],
    \label{eq:task_objective}
\end{equation}
where $C(\mathcal{X}_{\mathrm{final}})$ denotes the normalized
coverage of the settled garment, and the expectation is taken over
garment configurations and garment-dependent physical properties.

\subsection{Rotation-Induced Unfolding Dynamics}
We next explain how axial rotation produces the tension required to unfold a suspended garment. The Stage-I-selected grasp $\mathbf{g}^{\star}$ acts as an approximately fixed rotation anchor. Because the rotation axis is approximately vertical in our system, the radial distance of garment element $i$ from this axis is
\begin{equation}
    r_i
    =
    \left\|
        \mathbf{P}_{xy}
        \left(
            \mathbf{x}_i-\mathbf{g}^{\star}
        \right)
    \right\|_2,
    \label{eq:radial_distance}
\end{equation}
where $\mathbf{P}_{xy}$ projects a point onto the horizontal plane and the time index is omitted for clarity. Therefore, the selected grasp directly determines the radial distribution $\{r_i\}_{i=1}^{N}$ on which the rotation acts.

During rotation at angular velocity $\omega$, a locally co-rotating garment element experiences an outward centrifugal inertial load in the rotating reference frame:
\begin{equation}
    f_i^{\mathrm{cf}}
    =
    m_i\omega^2r_i,
    \label{eq:centrifugal_load}
\end{equation}
where $m_i$ is the mass of the element. Combining
\prettyref{eq:radial_distance} and
\prettyref{eq:centrifugal_load} shows the complementary roles of the grasp and angular velocity: the grasp determines the moment arm $r_i$, while the angular velocity controls the induced loading quadratically.


These distributed inertial loads are transmitted through the garment as
internal tension. Their cumulative effect beyond radial location $r$ is
approximated by the aggregate outward load
\begin{equation}
    F_{\mathrm{out}}(r)
    \approx
    \sum_{i:r_i\geq r} f_i^{\mathrm{cf}}
    =
    \omega^2
    \sum_{i:r_i\geq r} m_i r_i .
    \label{eq:radial_tension}
\end{equation}
where $r$ defines the radial threshold of the included garment elements.
The resulting internal tension straightens suspended regions and promotes
the separation of overlapping layers, explaining why axial rotation can
sustain an unfolding load without a large translational fling.
Note that \prettyref{eq:radial_tension} captures only the
rotation-induced loading; the full garment response also depends on
bending, damping and other factors.

The radial mass distribution can be summarized by the garment's moment of inertia about the grasp anchor:
\begin{equation}
    I_{\mathbf{g}^{\star}}
    =
    \sum_{i=1}^{N}
        m_i r_i^2.
    \label{eq:garment_inertia}
\end{equation}
A different grasp therefore changes both the element-wise moment arms and
the overall radial mass distribution.
Consequently, the same rotation command can produce substantially different unfolding behavior under different grasps or suspended configurations.

The complete mass distribution is unavailable from visual observations. We therefore approximate its radial spread using the observed garment point cloud:
\begin{equation}
    I_k^{\mathrm{geo}}
    =
    \frac{1}{N_k}
    \sum_{i=1}^{N_k}
    \left\|
        \mathbf{P}_{xy}
        \left(
            \mathbf{p}_{i,k}
            -
            \mathbf{g}^{\star}
        \right)
    \right\|_2^2,
    \label{eq:geometric_inertia}
\end{equation}
where $\{\mathbf{p}_{i,k}\}_{i=1}^{N_k}$ is the point cloud observed at rotation chunk $k$. The descriptor $I_k^{\mathrm{geo}}$ is an unweighted, observation-based proxy for radial spread rather than a physical estimate of
$I_{\mathbf{g}^{\star}}$. As the garment expands during rotation,
$I_k^{\mathrm{geo}}$ changes accordingly, providing Stage II with online information about the evolving suspended state.


This physical chain motivates the two-stage design. Stage I selects
the rotation anchor and thereby determines the initial radial mass
distribution. Stage II adapts the angular velocity and rotation
extent according to the evolving suspended garment state. The two
stages are coupled through the anchor
$\mathbf{g}^{\star}$ and the online radial-spread descriptor
$I_k^{\mathrm{geo}}$.

\subsection{Stage I: Rotation-Effective Grasp Ranking}
As indicated by \prettyref{eq:radial_distance}--\prettyref{eq:radial_tension},
the grasp point determines the rotation anchor and the resulting radial
distribution of the garment. A geometrically accessible grasp is
therefore not necessarily effective for rotation-induced unfolding.
As illustrated in \prettyref{fig:pipeline_overview}, Stage I learns to
select the visible grasp candidate expected to produce the highest
settled coverage.

From the initial RGB-D observation $O_0$, we reconstruct and preprocess
the visible garment point cloud
$\mathcal{P}_0=\{\mathbf{p}_i\}_{i=1}^{N_0}\subset\mathbb{R}^3$.
We then sample a
candidate set
\begin{equation}
    \mathcal{G}(O_0)
    =
    \{\mathbf{g}_j\}_{j=1}^{M}
    \subseteq
    \mathcal{P}_0,
    \qquad
    \left|\mathcal{G}(O_0)\right|
    =
    M,
    \label{eq:stage1_candidates}
\end{equation}
where $M$ denotes the number of candidates.
The sampling
procedure combines random visible points with high-point, boundary,
high-variation, and spatially distributed proposals to cover
geometrically distinct grasp regions.

For each candidate $\mathbf{g}_j\in\mathcal{G}(O_0)$, the scoring
network extracts a global garment feature $\mathbf{z}_0$, a local
point-cloud feature $\mathbf{h}_j$ around the candidate, and an explicit
geometric descriptor $\mathbf{f}_j$. These representations are fused to
predict the grasp score $s_j$, i.e.,
\begin{equation}
    \ell_j
    =
    H_{\phi}
    \left(
        \mathbf{z}_0,\mathbf{h}_j,\mathbf{f}_j
    \right),
    \qquad
    s_j
    =
    \sigma(\ell_j),
    \label{eq:stage1_scoring}
\end{equation}
where $H_{\phi}$ is the learned candidate-scoring model with parameters
$\phi$, $\ell_j$ is its output logit, and
$s_j\in(0,1)$ is the predicted grasp score. 

\begin{figure}[t]
    \centering
    \includegraphics[width=0.98\linewidth]{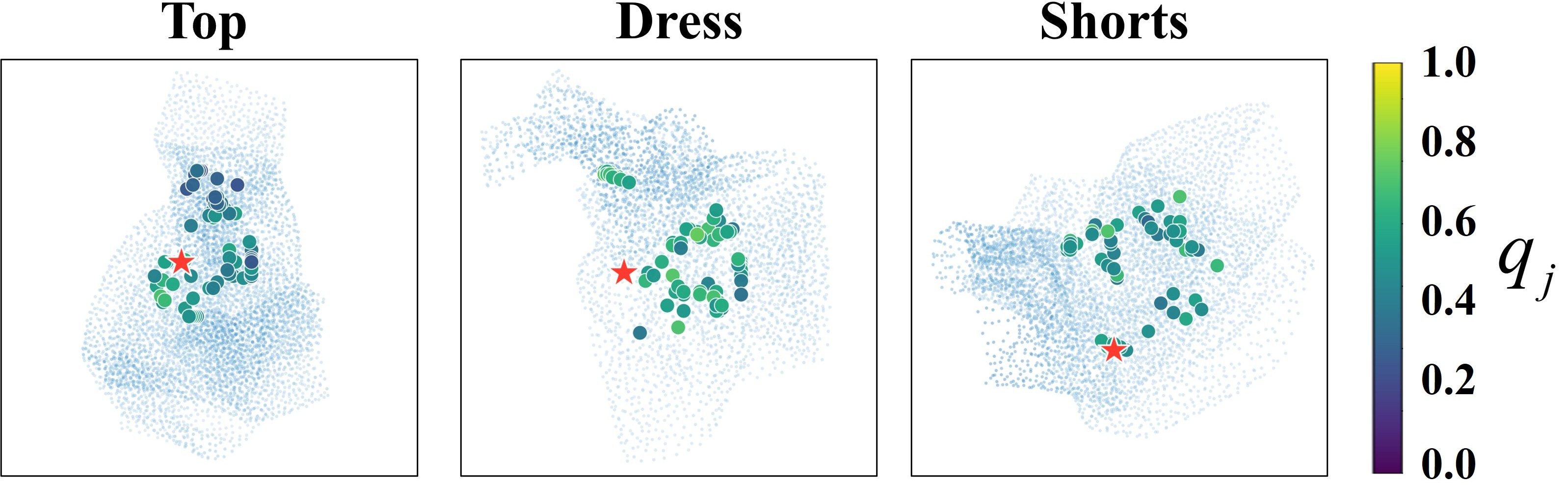}
    \vspace{-10pt}
    \caption{Stage I grasp-affordance labels of three simulated garment
    categories from ClothesNet \cite{zhou2023clothesnet}. Light-blue points show the observed garment point cloud, while colored markers denote sampled grasp candidates; color indicates the resulting settled coverage, and red stars mark the highest-quality candidates.}
    \label{fig:stageI_ex}
    \vspace{-15pt}
\end{figure}

The candidate-scoring model is trained using outcome supervision from
simulation. For each candidate $\mathbf{g}_j$, we execute the same
nominal rotation primitive $(T_0,\omega_0)$ and allow the garment to
settle into $\mathcal{X}_{\mathrm{final}}^{(j)}$. We denote the resulting
coverage by
$C_j=C(\mathcal{X}_{\mathrm{final}}^{(j)})$, using the same final-state
coverage measure as in \prettyref{eq:task_objective}. The corresponding
supervision label is
\begin{equation}
    q_j
    =
    \begin{cases}
        \min\left(C_j/C_{\max},\,1\right),
        & \text{successful grasp},\\
        0,
        & \text{failed grasp},
    \end{cases}
    \label{eq:stage1_rollout_label}
\end{equation}
where $C_{\max}$ is a fixed reference coverage used to scale successful
rollout outcomes to $[0,1]$; outcomes satisfying
$C_j\geq C_{\max}$ receive the maximum label $q_j=1$.
Examples of these rollout-derived labels are shown in
\prettyref{fig:stageI_ex}.

To learn both the absolute outcome and the relative ordering of
candidates, we combine a smooth-$L_1$ regression term with a pairwise
ranking term:
\begin{equation}
\begin{aligned}
    \mathcal{L}_{\mathrm I}
    ={}&
    \frac{1}{M}
    \sum_{j=1}^{M}
    \operatorname{SmoothL1}(s_j-q_j)\\
    &+
    \frac{\lambda_{\mathrm{rank}}}{|\mathcal{R}|}
    \sum_{(u,v)\in\mathcal{R}}
    \operatorname{softplus}
    \left[-(\ell_u-\ell_v)\right],
\end{aligned}
\label{eq:stage1_loss}
\end{equation}
where
$\mathcal{R}=\{(u,v)\mid q_u-q_v>\delta_q\}$ contains candidate pairs
separated by a quality margin $\delta_q$,
$\lambda_{\mathrm{rank}}$ weights the ranking term, and
$\operatorname{softplus}(x)=\log(1+\exp(x))$.

At inference, the highest-scoring candidate is selected without
executing additional rollouts:
\begin{equation}
\begin{aligned}
    j^{\star}
    &=
    \arg\max_{j\in\{1,\ldots,M\}} s_j,\\
    \mathbf{g}^{\star}
    &=
    \pi_{\mathrm I}(O_0)
    =
    \mathbf{g}_{j^{\star}},
    \qquad
    s^{\star}
    =
    s_{j^{\star}}.
\end{aligned}
\label{eq:stage1_selection}
\end{equation}
Therefore, \prettyref{eq:stage1_selection} instantiates the Stage-I
mapping defined in \prettyref{eq:two_stage_policies}. The selected grasp
$\mathbf{g}^{\star}$ becomes the rotation anchor, while its predicted
score $s^{\star}$ is passed to Stage II as part of the online garment
state.

\subsection{Stage II: Online Residual Rotation Control}


After grasping and lifting at $\mathbf g^\star$, Stage II retains the
same anchor and refines a nominal rotation primitive over $K$ online
control chunks.
At chunk $k$, the current garment point cloud
$\mathcal{P}_k=\{\mathbf{p}_{i,k}\}_{i=1}^{N_k}$ is summarized by a
compact observation. Let $\mathbf{c}_k$ denote its centroid and
$\mathbf{d}_k=\mathbf{c}_k-\mathbf{g}^{\star}$ its displacement from
the grasp anchor. The complete observation is
\begin{equation}
\begin{aligned}
    \mathbf{o}_k
    =
    [\,
    &\mathbf{d}_k,\,
    \|\mathbf{d}_k\|_2,\,
    I_k^{\mathrm{geo}},\,
    \sigma_{z,k},\,
    s^{\star},\\
    &\rho_k,\,
    \gamma_k,\,
    \bar{k},\,
    n_k,\,
    \omega_0,\,
    v_{\mathrm{ref}}
    \,]
    \in
    \mathbb{R}^{13}.
\end{aligned}
\label{eq:stage2_observation}
\end{equation}
Here, $I_k^{\mathrm{geo}}$ is the grasp-centered radial-spread
descriptor defined in \prettyref{eq:geometric_inertia}, and
$\sigma_{z,k}$ is the vertical standard deviation of the garment
points. Together, they provide compact geometric cues about the
suspended garment. The score $s^{\star}$, given in
\prettyref{eq:stage1_selection}, supplies the Stage-I grasp-quality
prior.

The remaining features describe unfolding and motion progress.
Specifically, $\rho_k=A_k/A_{\mathrm{ref}}$ is the current covered-area
ratio relative to the fully unfolded reference, while
$\gamma_k=A_k/A_{k-1}$ measures the improvement relative to the
preceding chunk. The variables $\bar{k}$ and $n_k$ denote the normalized
chunk index and accumulated rotation in turns, respectively.
The nominal angular velocity $\omega_0$ provides the reference motion
context, and $v_{\mathrm{ref}}\in\{0,1\}$ indicates whether the reference
area $A_{\mathrm{ref}}$ is available. Therefore,
\prettyref{eq:stage2_observation} connects the Stage-I grasp decision,
the current garment geometry, and the progress of the ongoing rotation.

Following the Stage-II mapping as in
\prettyref{eq:two_stage_policies}, the policy $\pi_{\mathrm{II}}$ produces a two-dimensional normalized residual action:
\begin{equation}
\begin{aligned}
    \mathbf{a}_k
    &=
    \begin{bmatrix}
        a_k^T & a_k^\omega
    \end{bmatrix}^{\mathsf T}
    =
    \pi_{\mathrm{II}}(\mathbf{o}_k)
    \in[-1,1]^2,\\
    \Delta T_k
    &=
    \alpha_T a_k^T,
    \qquad
    \Delta\omega_k
    =
    \alpha_\omega a_k^\omega.
\end{aligned}
\label{eq:stage2_action}
\end{equation}
The two outputs adjust \textit{how far} and \textit{how fast} the next rotation chunk is
executed. The scaling factors $\alpha_T$ and $\alpha_\omega$ convert the
normalized policy outputs into residual rotation extent and angular
velocity, respectively.

The executed command is obtained by adding these residuals to the
nominal primitive:
\begin{equation}
\begin{aligned}
    T_k
    &=
    \operatorname{clip}
    \left(
        T_0+\Delta T_k,\,
        T_{\min},T_{\max}
    \right),\\
    \omega_k
    &=
    \operatorname{clip}
    \left(
        \omega_0+\Delta\omega_k,\,
        \omega_{\min},\omega_{\max}
    \right).
\end{aligned}
\label{eq:stage2_residual_control}
\end{equation}
Here, $T_0$ and $T_k$ are the nominal and executed state-adaptive rotation
extents in turns. Similarly, $\omega_0$ and $\omega_k$ are the corresponding angular
velocities in $\mathrm{rad/s}$. The $\operatorname{clip}$ operation in \prettyref{eq:stage2_residual_control}
simply limits each command to the feasible interval. 

After executing chunk $k$, the garment is observed again and
$\mathbf{o}_{k+1}$ is constructed under the same grasp. This produces
online state-conditioned adaptation while avoiding regrasping within
an attempt. The duration of each chunk and the cumulative release time
are
\begin{equation}
    \tau_k
    =
    \frac{2\pi T_k}{|\omega_k|},
    \qquad
    t_{\mathrm{rel}}
    =
    \sum_{k=0}^{K-1}\tau_k.
    \label{eq:stage2_release_timing}
\end{equation}
The gripper opens after the final chunk. Although the number of chunks
is fixed, \prettyref{eq:stage2_release_timing} shows that the physical
release timing is adaptive because it is jointly determined by the
state-conditioned rotation extents and angular velocities. Thus, the
policy determines when to release through the learned rotation
schedule.

We train $\pi_{\mathrm{II}}$ using proximal policy optimization
(PPO)~\cite{schulman2017proximal}. The per-chunk reward combines bounded
coverage improvement with action and boundary penalties:
\begin{equation}
\begin{aligned}
    r_k
    ={}&
    w_{\Delta}\widehat{\Delta C}_k
    -
    w_a\|\mathbf{a}_k\|_2^2
    -
    w_b B_k\\
    &+
    \mathbf{1}_{\{k=K-1\}}R_{\mathrm{term}},
\end{aligned}
\label{eq:stage2_reward}
\end{equation}
where $\widehat{\Delta C}_k$ is the bounded coverage change produced by
the current chunk, $B_k$ penalizes command values that reach or exceed
the feasible control boundaries, and $w_{\Delta}$, $w_a$, and $w_b$
are their respective weights. The terminal reward $R_{\mathrm{term}}$ is added after the final
rotation chunk and it evaluates the coverage immediately before release
and after settling, together with the corresponding improvements.

Together, \prettyref{eq:stage2_observation}--\prettyref{eq:stage2_release_timing} instantiate the Stage-II policy
defined in \prettyref{eq:two_stage_policies}. 
In summary, Stage I determines where the rotation should be anchored, while Stage II determines how fast and how far to rotate and, consequently, when to release. This coupling enables \textit{RotateIt!} to adapt a compact dynamic primitive to diverse garment configurations without learning an unconstrained robot trajectory.





\begin{table*}[t]
    \centering
    \caption{Per-garment unfolding results. $S@k$ denotes success within
    $k$ attempts; Overall aggregates all rollouts in each scenario.}
    \vspace{-10pt}
    \label{tab:overall_performance}
    \setlength{\tabcolsep}{1.5pt}
    \renewcommand{\arraystretch}{1.08}

    \begin{threeparttable}
    \scalebox{0.94}{%
    \begin{tabular}{ll*{21}{c}}
        \toprule
        \multirow{2}{*}{Metric}
        & \multirow{2}{*}{Method}
        & \multicolumn{6}{c}{Seen Simulation}
        & \multicolumn{6}{c}{Unseen Simulation}
        & \multicolumn{9}{c}{Unseen Real World} \\
        \cmidrule(lr){3-8}
        \cmidrule(lr){9-14}
        \cmidrule(lr){15-23}

        &
        & S1 & S2 & S3 & S4 & S5 & Overall
        & U1 & U2 & U3 & U4 & U5 & Overall
        & R1 & R2 & R3 & R4
        & R5 & R6 & R7 & R8 & Overall \\
        \midrule

        \multirow{2}{*}{Cov.@1 $\uparrow$}
        & P\&P
        & 0.609 & 0.629 & 0.579 & 0.667 & 0.574 & 0.612
        & \textbf{0.726} & 0.625 & 0.621 & 0.771 & 0.623 & 0.673
        & 0.657 & 0.555 & 0.434 & 0.514
        & 0.497 & 0.530 & 0.420 & 0.431 & 0.505 \\
        & Ours
        & \textbf{0.739} & \textbf{0.778} & \textbf{0.709}
        & \textbf{0.867} & \textbf{0.680} & \textbf{0.755}
        & 0.717 & \textbf{0.786} & \textbf{0.704}
        & \textbf{0.793} & \textbf{0.724} & \textbf{0.745}
        & \textbf{0.793} & \textbf{0.825} & \textbf{0.640}
        & \textbf{0.684} & \textbf{0.722} & \textbf{0.751}
        & \textbf{0.624} & \textbf{0.656} & \textbf{0.712} \\
        \addlinespace[1pt]

        \multirow{2}{*}{Final Cov. $\uparrow$}
        & P\&P
        & 0.601 & 0.638 & 0.585 & 0.701 & 0.570 & 0.619
        & 0.799 & 0.679 & 0.658 & 0.834 & 0.653 & 0.725
        & 0.779 & 0.747 & 0.606 & 0.654
        & 0.744 & 0.738 & 0.619 & 0.584 & 0.684 \\
        & Ours
        & \textbf{0.820} & \textbf{0.879} & \textbf{0.800}
        & \textbf{0.962} & \textbf{0.756} & \textbf{0.843}
        & \textbf{0.884} & \textbf{0.910} & \textbf{0.767}
        & \textbf{0.872} & \textbf{0.868} & \textbf{0.860}
        & \textbf{0.927} & \textbf{0.906} & \textbf{0.811}
        & \textbf{0.872} & \textbf{0.923} & \textbf{0.902}
        & \textbf{0.778} & \textbf{0.765} & \textbf{0.860} \\
        \addlinespace[1pt]

        \multirow{2}{*}{S@1 (\%) $\uparrow$}
        & P\&P
        & 0.0 & 5.0 & 0.0 & 0.0 & 0.0 & 1.0
        & 20.0 & 10.0 & 5.0 & 40.0 & 5.0 & 16.0
        & 15.0 & 10.0 & 0.0 & 5.0
        & 0.0 & 0.0 & 0.0 & 0.0 & 3.8 \\
        & Ours
        & \textbf{15.0} & \textbf{30.0} & \textbf{25.0}
        & \textbf{60.0} & \textbf{10.0} & \textbf{28.0}
        & \textbf{25.0} & \textbf{55.0} & \textbf{20.0}
        & \textbf{50.0} & \textbf{25.0} & \textbf{35.0}
        & \textbf{50.0} & \textbf{50.0} & \textbf{40.0}
        & \textbf{35.0} & \textbf{35.0} & \textbf{50.0}
        & \textbf{5.0} & \textbf{35.0} & \textbf{37.5} \\
        \addlinespace[1pt]

        \multirow{2}{*}{S@3 (\%) $\uparrow$}
        & P\&P
        & 0.0 & 5.0 & 5.0 & 15.0 & 0.0 & 5.0
        & 65.0 & 20.0 & 10.0 & 70.0 & 10.0 & 35.0
        & 45.0 & 45.0 & 15.0 & 40.0
        & 40.0 & 35.0 & 15.0 & 15.0 & 31.2 \\
        & Ours
        & \textbf{75.0} & \textbf{70.0} & \textbf{55.0}
        & \textbf{95.0} & \textbf{35.0} & \textbf{66.0}
        & \textbf{80.0} & \textbf{95.0} & \textbf{60.0}
        & \textbf{80.0} & \textbf{80.0} & \textbf{79.0}
        & \textbf{90.0} & \textbf{80.0} & \textbf{65.0}
        & \textbf{80.0} & \textbf{85.0} & \textbf{90.0}
        & \textbf{60.0} & \textbf{55.0} & \textbf{75.6} \\
        \addlinespace[1pt]

        \multirow{2}{*}{Attempts $\downarrow$}
        & P\&P
        & 3.00 & 2.90 & 2.95 & 2.95 & 3.00 & 2.96
        & 2.25 & 2.75 & 2.90 & 2.00 & 2.85 & 2.55
        & 2.10 & 2.60 & 2.70 & 2.60
        & 2.85 & 2.75 & 2.95 & 2.95 & 2.69 \\
        & Ours
        & \textbf{2.30} & \textbf{2.05} & \textbf{2.35}
        & \textbf{1.50} & \textbf{2.55} & \textbf{2.15}
        & \textbf{2.15} & \textbf{1.60} & \textbf{2.40}
        & \textbf{1.80} & \textbf{2.15} & \textbf{2.02}
        & \textbf{1.70} & \textbf{1.80} & \textbf{1.95}
        & \textbf{1.85} & \textbf{1.95} & \textbf{1.75}
        & \textbf{2.35} & \textbf{2.10} & \textbf{1.93} \\
        \bottomrule
    \end{tabular}%
     }

    \begin{tablenotes}[flushleft]
        \scriptsize
        \item[]%
        \parbox[t]{0.98\textwidth}{%
            Note: 
            S1--S5: Shorts-1, Shorts-2, Dress-1, Dress-2, and Top;
            U1--U5: Dress-1, Dress-2, Shorts, Pants, and Top;
            R1--R8: See right portion of \prettyref{fig:exp_result_sim2real_R1toR8}.
            Best results are shown in bold.%
        }
    \end{tablenotes}
    \end{threeparttable}
    \vspace{-15pt}
\end{table*}

\section{Experiments}
\label{sec:exp}
We conduct simulation and real-world experiments to answer four questions. \textbf{Q1:} How does \textit{RotateIt!} compare with quasi-static P\&P in unfolding effectiveness and action efficiency? \textbf{Q2:} How do the two learned stages and the underlying rotation dynamics contribute to its performance? \textbf{Q3:} Does the simulation-trained policy generalize to unseen simulated and real garments without real-world fine-tuning? \textbf{Q4:} Can the unfolded states produced by \textit{RotateIt!} support downstream folding?

\subsection{Experimental Setup and Evaluation Metrics}
\label{sec:experimental_setup}

We conduct simulation experiments in Isaac Sim using a Franka robot. Five simulated garments are used during training, and performance is evaluated on new configurations of these seen garments and five unseen simulated garments. Real-world evaluation uses a single-arm robot (xArm 7), two RGB-D cameras (RealSense D435 and D405), and eight unseen garments. Both learned stages are trained entirely in simulation and transferred to the real system without fine-tuning.
To promote generalization, training covers variations in garment size, mass, friction, stiffness, and damping. Random initial configurations are generated by varying the garment's three-dimensional orientation and drop height. Within each \textit{RotateIt!} attempt, Stage II executes $K=4$ online control chunks under the same grasp following \prettyref{eq:stage2_residual_control}.

We conduct 20 rollouts for every garment--method pair. A rollout is considered successful when the coverage reaches $0.8$ and terminates upon success or after three attempts. Following an unsuccessful attempt, the garment is allowed to settle and is observed again before the next action. Coverage is computed as
$C(\mathcal{X})=A(\mathcal{X})/A_{\mathrm{ref}}$, where $A(\mathcal{X})$ is the observed garment area and $A_{\mathrm{ref}}$ is its fully spread reference area. For each real garment, $A_{\mathrm{ref}}$ is acquired beforehand. The coverage is reported without clipping and may slightly exceed one because of observation variations. We report the mean coverage after the first attempt (Cov.@1), coverage at rollout termination (Final Cov.), cumulative success rate within $k$ attempts (S@$k$), and the average number of executed attempts (Attempts). Here, Attempts serves as the action-efficiency measure, since each additional attempt requires another manipulation, settling period, and observation.

\subsection{Comparison with Quasi-Static Unfolding}
\label{subsec:comparison}

\textbf{Baseline and protocol.}
This subsection answers \textbf{Q1}: whether dynamic axial rotation provides more effective unfolding than quasi-static manipulation. We compare \textit{RotateIt!} with a quasi-static pick-and-place (P\&P) baseline \cite{lee2021learning}. P\&P selects the highest-valued feasible grasp from learned spatial action-value maps, with the corresponding discrete direction and distance specifying the placement endpoint. It is trained in the same simulation environment, on the same five seen garments, and under the same initialization and physical variations as \textit{RotateIt!}. Under the common protocol in \prettyref{sec:experimental_setup}, one P\&P attempt executes one pick-and-place action, whereas one \textit{RotateIt!} attempt executes one grasp--rotate--release action.

\textbf{Unfolding effectiveness.}
As reported in \prettyref{tab:overall_performance}, \textit{RotateIt!} improves Cov.@1 over P\&P by $+14.3$, $+7.2$, and $+20.7$ percentage points on seen simulation, unseen simulation, and unseen real garments, respectively, and achieves higher first-attempt coverage on 17 of the 18 garments. In real-world tests, 
\textit{RotateIt!} improves the Cov.@1 from $0.505$ to $0.712$, corresponding to a $+41\%$ relative gain over P\&P. This larger progress per attempt also raises S@1 by $+19.0$ to $+33.7$ percentage points across the three settings.


The advantage persists over complete rollouts. \textit{RotateIt!} achieves S@3 values of $66.0\%$, $79.0\%$, and $75.6\%$ on seen simulation, unseen simulation, and unseen real garments, exceeding P\&P by $+61.0$, $+44.0$, and $+44.4$ percentage points, respectively. Final Cov. increases by $+22.4$, $+13.5$, and $+17.6$ percentage points and is higher on all 18 garments. In contrast, P\&P produces \textit{no} single-attempt successes on four of the five seen simulated garments and five of the eight real garments. Even after three attempts, it achieves \textit{no} successful rollout on S1 or S5 and fails in $95.0\%$, $65.0\%$, and $68.8\%$ of the rollouts across three evaluation scenarios. 

In addition, \textit{RotateIt!} requires only $2.15$, $2.02$, and $1.93$ attempts on average, compared with $2.96$, $2.55$, and $2.69$ for P\&P, corresponding to reductions of $27\%$, $21\%$, and $28\%$. By avoiding repeated manipulation, settling, and re-observation cycles, \textit{RotateIt!} reaches successfully unfolded states more rapidly while also achieving higher terminal coverage.

\begin{figure*}[thbp]
    \centering
    \includegraphics[width=0.998\textwidth]{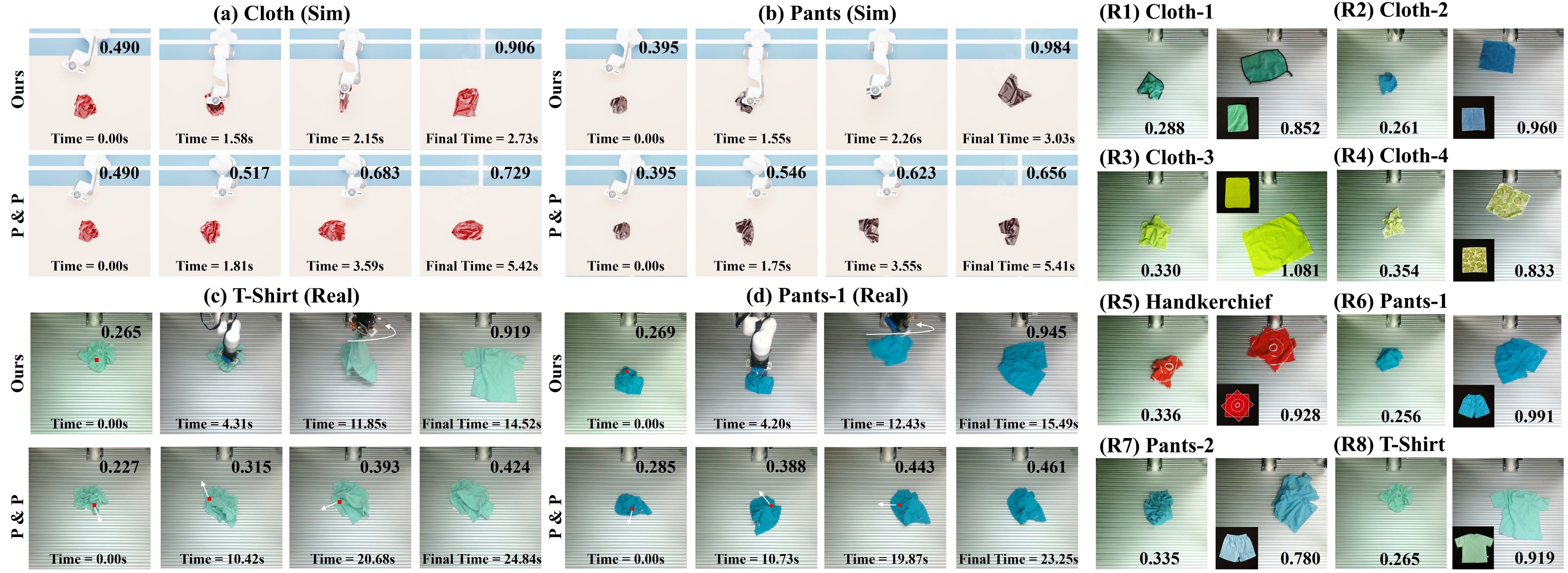}
    \vspace{-20pt}
    \caption{Representative unfolding results. Left: comparison with
    quasi-static P\&P in simulation (a)--(b) and the real world (c)--(d);
    \textit{RotateIt!} achieves higher coverage with one
    grasp--rotate--release attempt, whereas P\&P uses sequential actions.
    Right: zero-shot results on eight unseen real garments (R1--R8), showing
    the initial and final settled states; insets show the fully spread
    references. Numbers report coverage, and timestamps are shown for the
    comparison rollouts.}
    \label{fig:exp_result_sim2real_R1toR8}
    \vspace{-10pt}
\end{figure*}

\textbf{Qualitative analysis.}
The representative rollouts in the left portion of \prettyref{fig:exp_result_sim2real_R1toR8} illustrate the source of this advantage. P\&P produces localized changes through successive actions, whereas rotation-induced tension separates overlapping layers and generates a large, nonlocal deformation within one attempt. Thus, \textit{RotateIt!} can replace multiple incremental pick-and-place actions with a single dynamic manipulation primitive. Additional rollouts and representative failure cases are provided in the supplementary video.


\begin{figure}[t]
    \centering
    \includegraphics[width=0.98\linewidth]{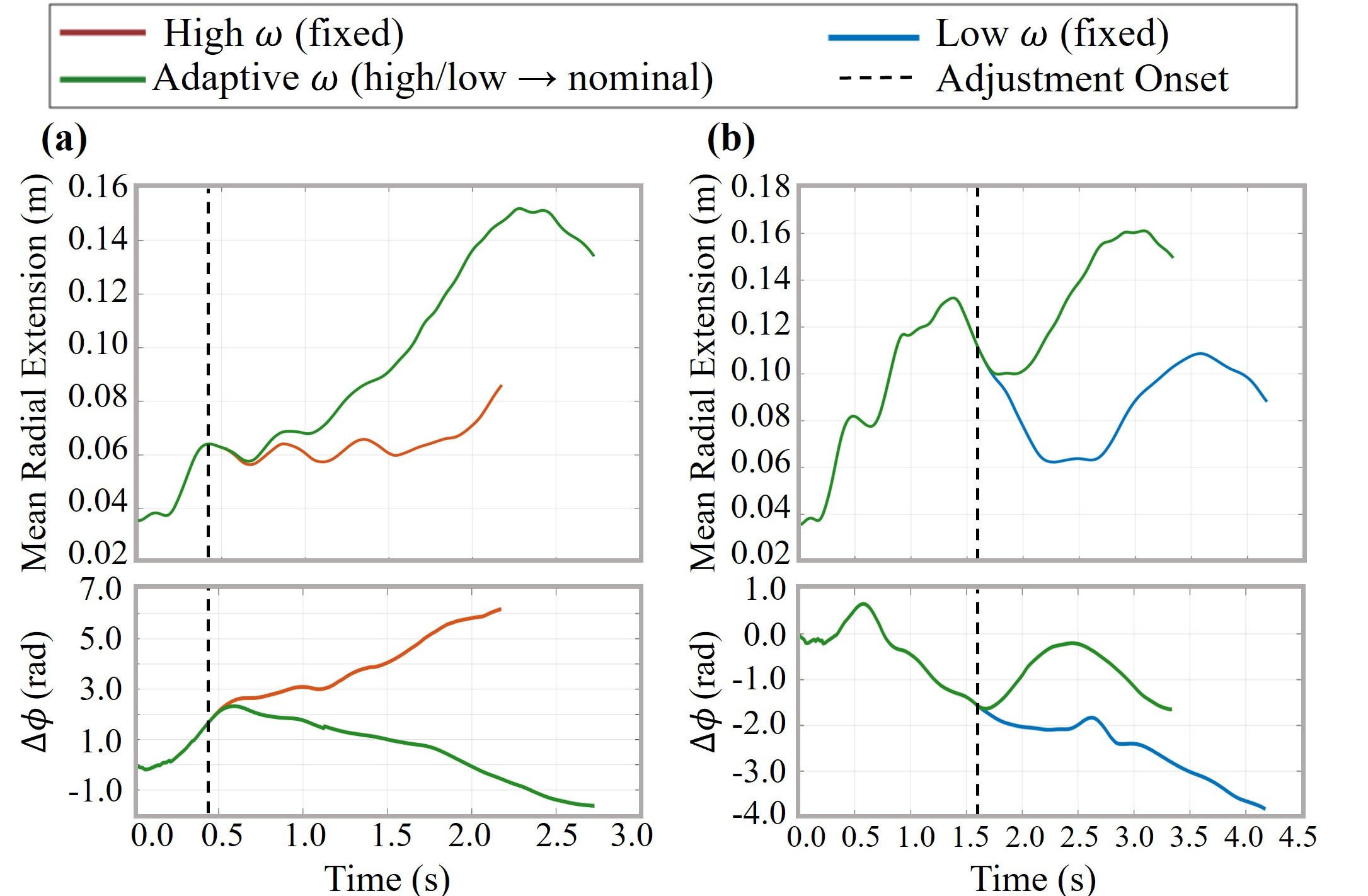}
    \vspace{-10pt}
    \caption{Controlled analysis of angular-velocity adaptation from high
    (a) and low (b) initial speeds. Adjusting $\omega$ toward the nominal
    $\omega_0$ increases radial extension and suppresses distal phase
    deviation $\Delta\phi$ relative to fixed-speed execution; dashed lines
    mark the adjustment onset.}
    \label{fig:rotation_dynamics}
    \vspace{-10pt}
\end{figure}

\subsection{Component Ablations and Rotation-Dynamics Analysis}
\label{subsec:ablation_dynamics}
\textbf{Component ablations.}
This subsection answers \textbf{Q2}: which components enable effective and reliable rotational unfolding. \prettyref{tab:ablation} reports a separate ablation evaluation using independently resampled seen-simulation rollouts.
The full model achieves a Cov.@1 of $0.743$, an S@3 of $70.0\%$, and a Final Cov. of $0.835$. Replacing the learned grasp ranker with edge-random sampling (M1) reduces these metrics by $16.3$, $52.0$, and $20.1$ percentage points, respectively, confirming that a geometrically feasible grasp is not necessarily effective for rotational unfolding. Fixing both Stage-II parameters at their nominal values $(T_0,\omega_0)$ (M2) similarly reduces Cov.@1 and S@3 by $9.3$ and $38.0$ percentage points, respectively, demonstrating the need for observation-conditioned control.

With the learned grasp retained, M3 and M4 isolate speed and extent adaptation, respectively. Adaptive extent (M4) provides a larger improvement in immediate coverage than adaptive speed alone (M3), indicating that selecting an appropriate rotation extent primarily determines how much deformation is produced. Nevertheless, adding speed adaptation to M4 increases S@3 from $54.0\%$ to $70.0\%$, while Cov.@1 increases more moderately from $0.722$ to $0.743$. Thus, speed adaptation contributes particularly to rollout-level reliability, and the two control dimensions are complementary rather than interchangeable.

\textbf{Rotation-dynamics analysis.}
%
\prettyref{fig:rotation_dynamics} provides  evidence for why a fixed angular velocity is insufficient. The mean radial extension measures the outward spread around the grasp, while the distal phase deviation $\Delta\phi$ characterizes the angular mismatch between the grasp motion and the garment's distal response. At a fixed high speed, $\Delta\phi$ grows continuously while radial extension remains limited, indicating that the distal garment cannot coherently follow the imposed rotation. At a fixed low speed, the garment exhibits pendulum-like overshoot: extension increases transiently but subsequently contracts and oscillates. Adapting the angular velocity from either initial setting toward the
nominal $\omega_0$ interrupts these phase-divergent behaviors and produces greater, more sustained radial extension.

These observations complement the aggregate-loading relation in \prettyref{eq:radial_tension}: although rotation-induced loading increases with angular velocity, higher speed does not necessarily yield better unfolding because tension must propagate through a compliant garment. Note that the adjustment onset in \prettyref{fig:rotation_dynamics} is introduced only for this controlled diagnostic and is not a trigger used by the deployed policy. 
Together, the ablation and dynamics results support the two-stage design: Stage I selects an anchor that induces a favorable suspended geometry, while Stage II adapts the motion to the resulting garment dynamics.

\begin{table}[t]
    \centering
    \caption{Component Ablations on Seen-Simulation Rollouts.}
    \vspace{-10pt}
    \label{tab:ablation}
    \begin{threeparttable}
        \scalebox{0.74}{%
            \begin{tabular}{llcccccc}
                \toprule
                \multirow{2}{*}{Method}
                & \multicolumn{1}{c}{Ablation from}
                & \multicolumn{1}{c}{Stage I}
                & \multicolumn{2}{c}{Stage II}
                & \multicolumn{3}{c}{Performance $\uparrow$} \\
                \cmidrule(lr){2-2}
                \cmidrule(lr){3-3}
                \cmidrule(lr){4-5}
                \cmidrule(lr){6-8}

                & \textit{RotateIt!}
                & Grasp
                & Extent
                & Speed
                & Cov.@1 
                & S@3 (\%) 
                & Final Cov. \\
                \midrule

                M1
                & Random Grasp
                & $\sim$
                & $\checkmark$
                & $\checkmark$
                & 0.580
                & 18.0
                & 0.634 \\

                M2
                & Fixed Stage II
                & $\checkmark$
                & $-$
                & $-$
                & 0.650
                & 32.0
                & 0.692 \\

                M3
                & Fixed Extent
                & $\checkmark$
                & $-$
                & $\checkmark$
                & 0.650
                & 40.0
                & 0.725 \\

                M4
                & Fixed Speed
                & $\checkmark$
                & $\checkmark$
                & $-$
                & 0.722
                & 54.0
                & 0.761 \\

                Ours
                & Full Model
                & $\checkmark$
                & $\checkmark$
                & $\checkmark$
                & \textbf{0.743}
                & \textbf{70.0}
                & \textbf{0.835} \\

                \bottomrule
            \end{tabular}%
        }

        \begin{tablenotes}[flushleft]
            \scriptsize
            \item[]%
            \parbox[t]{0.98\columnwidth}{%
                Note: $\checkmark$, $\sim$, and $-$ denote learned/adaptive,
                random, and fixed-at-nominal components,
                respectively. The nominal extent and speed are
                $(T_0,\omega_0)$.%
            }
        \end{tablenotes}
    \end{threeparttable}
    \vspace{-20pt}
\end{table}

\subsection{Generalization and Zero-Shot Sim-to-Real Transfer}
\label{subsec:generalization}
This subsection answers \textbf{Q3}. Both stages of \textit{RotateIt!}
are trained exclusively in simulation. The resulting policies are
deployed unchanged to unseen simulated and real garments,
without real-world fine-tuning or garment-specific retraining.


\textbf{Unseen-garment generalization.}
As shown in \prettyref{tab:overall_performance}, performance remains stable when moving from seen to unseen simulated garments. Cov.@1 decreases by only $1.0$ percentage point, from $0.755$ to $0.745$, while Final Cov. increases from $0.843$ to $0.860$ and S@3 rises from $66.0\%$ to $79.0\%$. These results indicate that the learned grasp ranking and residual rotation control are not tied to the garment identities used during training.


\textbf{Zero-shot real-world transfer.}
Without real-world fine-tuning, \textit{RotateIt!} achieves a Cov.@1 of $0.712$, a Final Cov. of $0.860$, and an S@3 of $75.6\%$ across eight unseen real garments. Relative to unseen simulation, the real-world results exhibit only a $3.3$-point reduction in Cov.@1 and a $3.4$-point reduction in S@3, while retaining the same Final Cov. This small transfer gap provides strong evidence that the simulation-trained policies capture garment-state and rotational-dynamics cues that remain effective under real sensing and physical interactions.


\textbf{Qualitative analysis.}
The qualitative results in the right portion of \prettyref{fig:exp_result_sim2real_R1toR8} further demonstrate transfer across diverse garment geometries, sizes, and materials. According to \prettyref{tab:overall_performance}, four garments (R1, R2, R5, and R6) achieve average Final Cov. above $0.90$, with R1 and R6 each reaching an S@3 of $90.0\%$. Performance is lower on the more strongly branched Pants-2 (R7) and T-Shirt (R8), whose extended parts can remain folded or form self-contact during suspension, making the strict $0.9$ threshold harder to reach. Nevertheless, their S@3 values remain $60.0\%$ and $55.0\%$, respectively. 

Overall, the small performance variation between seen and unseen
simulation garments, together with unchanged deployment across eight
unseen real garments, demonstrates strong garment-level generalization
and zero-shot sim-to-real transfer. The supplementary video provides
continuous unedited trials in which the garments are repeatedly
re-crumpled and presented to the robot in new initial states.

\subsection{Downstream Folding Application}
\label{subsec:downstream_folding}

\begin{figure}[t]
    \centering
    \includegraphics[width=0.98\linewidth]{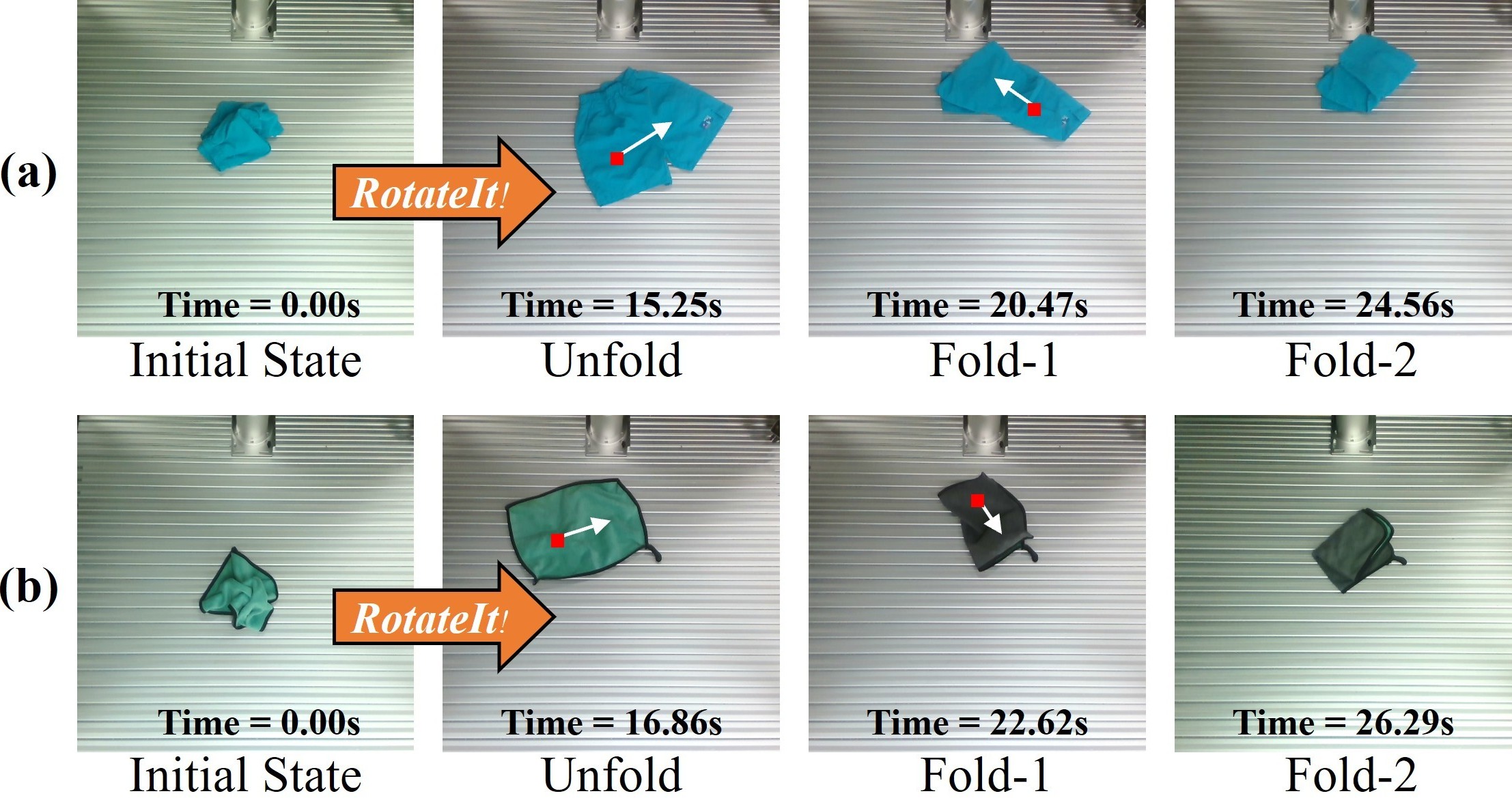}
    \vspace{-10pt}
    \caption{Downstream folding on two unseen real garments. Starting from a randomly crumpled configuration, \textit{RotateIt!} first unfolds each garment, and a two-step folding policy subsequently operates directly on the settled result without manual rearrangement. Red squares and white arrows indicate the grasp points and motion directions, respectively.}
    \label{fig:downstreaming_tasks}
    \vspace{-15pt}
\end{figure}

This subsection answers \textbf{Q4}: whether the states produced by \textit{RotateIt!} can serve directly as task-ready inputs for downstream garment manipulation. Although \textit{RotateIt!} is trained only for unfolding, we directly connect its output to a keypoint-based two-step folding policy and evaluate the complete pipeline on two unseen real garments. From each settled unfolding result, the folding policy automatically determines the grasp and placement locations from garment keypoints and executes two sequential pick-and-place actions. As shown in \prettyref{fig:downstreaming_tasks}, the robot transforms each randomly crumpled garment into a folded configuration without manual rearrangement, intermediate state reset, or human intervention. The complete autonomous unfold-to-fold sequences finish in $24.56$\,s and $26.29$\,s, respectively. Within comparable execution times, the representative P\&P rollouts in the left portion of \prettyref{fig:exp_result_sim2real_R1toR8} still fail to produce task-ready unfolded states, whereas \textit{RotateIt!} has already completed both downstream folding actions. These demonstrations show that its zero-shot real-world outputs are not merely high-coverage states, but directly actionable intermediate configurations that enable fast and fully autonomous unfold-to-fold manipulation, suggesting the potential of \textit{RotateIt!} as a general-purpose initialization primitive for broader downstream garment manipulation tasks.


\section{Conclusion and future work}
\label{sec:conclusion}

We presented \textit{RotateIt!}, a single-arm dynamic garment unfolding framework that uses adaptive axial rotation to achieve higher coverage with fewer attempts. Its two-stage policy couples rotation-effective grasp selection with online residual control of the rotation speed and extent, thereby adapting the release timing to the evolving suspended garment state. Experiments on seen and unseen simulated garments and eight unseen real garments demonstrate higher coverage and success rates with fewer attempts than the quasi-static pick-and-place method. Notably, the simulation-trained policy transfers directly to unseen real garments without real-world fine-tuning or garment-specific adaptation. Downstream folding further demonstrates the utility of the resulting unfolded states. Future work will extend \textit{RotateIt!} to goal-conditioned garment configurations and environment-aware manipulation that accounts for surrounding geometry and garment--environment interactions.

\bibliographystyle{IEEEtran}
\typeout{}
\bibliography{IEEEabrv,mybibfiles}
\end{document}